\documentclass[letterpaper]{article} 
\usepackage{aaai2027} 
\usepackage[hyphens]{url} 
\usepackage{graphicx} 
\usepackage{natbib} 
\usepackage{caption} 
\usepackage{booktabs}
\usepackage{amsmath}
\usepackage{eso-pic}
\title{LongDocBench: Benchmarking TOC Hierarchy and Contextual Relationship Recovery in Long Documents}

\author{
    Yuefeng Zou\textsuperscript{1}\equalcontrib,
    Yichen Lu\textsuperscript{1}\equalcontrib,
    Jingxiao Yang\textsuperscript{1,2}\equalcontrib,
    Bingtao Fu\textsuperscript{1},\\
    Gaoyang Zhang\textsuperscript{1},
    Xiongfei Bai\textsuperscript{1},
    Tian Chen\textsuperscript{1},
    Xiang Qi\textsuperscript{1}
}
\affiliations{
    \textsuperscript{1}Ant Group \\
    \textsuperscript{2}Zhejiang University
}

\begin{document}

\AddToShipoutPictureFG*{%
    \AtPageUpperLeft{%
        \raisebox{-1.8cm}[0pt][0pt]{%
            \hspace*{1cm}%
            \begin{minipage}{\dimexpr\paperwidth-2cm\relax}
                \begin{minipage}[c]{0.07\paperwidth}
                    \includegraphics[width=\linewidth]{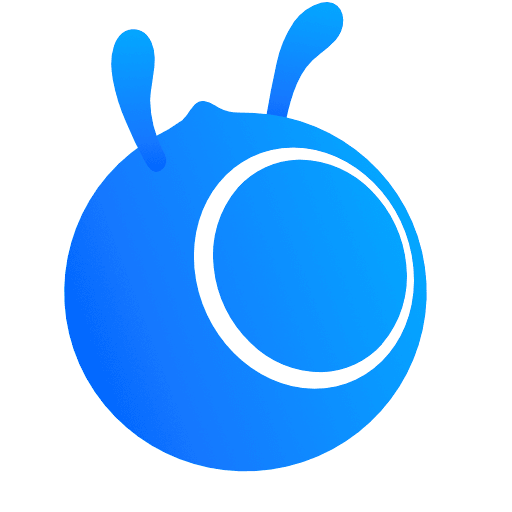}
                \end{minipage}
                \begin{minipage}[c]{0.25\paperwidth}
                    {\sffamily\bfseries\Large ANT \\[0.1em] GROUP}
                \end{minipage}


                \noindent\rule{\linewidth}{0.5pt}
            \end{minipage}%
        }%
    }%
}

\maketitle

\begin{abstract}
Parsing visual documents into machine-readable representations is fundamental to document intelligence. Existing benchmarks focus on page-level element recognition, reading order, formula recognition, and table structure. Long documents, however, also require document-level structure recovery. This includes reconstructing cross-page table-of-contents (TOC) hierarchies and identifying typed links from tables and figures to their captions, notes, and sources, often in one-to-many form. Because these structures are covered only partially or subsumed within broader parsing protocols, existing benchmarks cannot directly evaluate two key document-level tasks: \emph{Table-of-Contents Hierarchy Recovery} and \emph{Contextual Relationship Recovery}.
To benchmark these two tasks, we introduce \textsc{LongDocBench}, comprising 85 real-world financial reports, textbooks, and academic papers spanning 2,582 pages, with up to 105 pages per document. It provides human-verified annotations for 3,937 heading nodes (mean node depth 3.55; maximum depth 9) and 3,258 contextual relationships annotated across 2,680 table and figure objects.
We further evaluate both the downstream utility and recoverability of these structures. Long-document question-answering experiments show that human-verified TOC hierarchies and contextual relationships improve reasoning, with their combination providing complementary benefits. Meanwhile, representative document parsers remain limited on both recovery tasks despite strong page-level performance. To support further progress, we publicly release \textsc{LongDocBench} and its evaluation protocol and reproducible testbed for advancing document-level structure recovery in long documents.
\end{abstract}

\begin{figure}[t]
    \centering
    \includegraphics[width=\columnwidth]{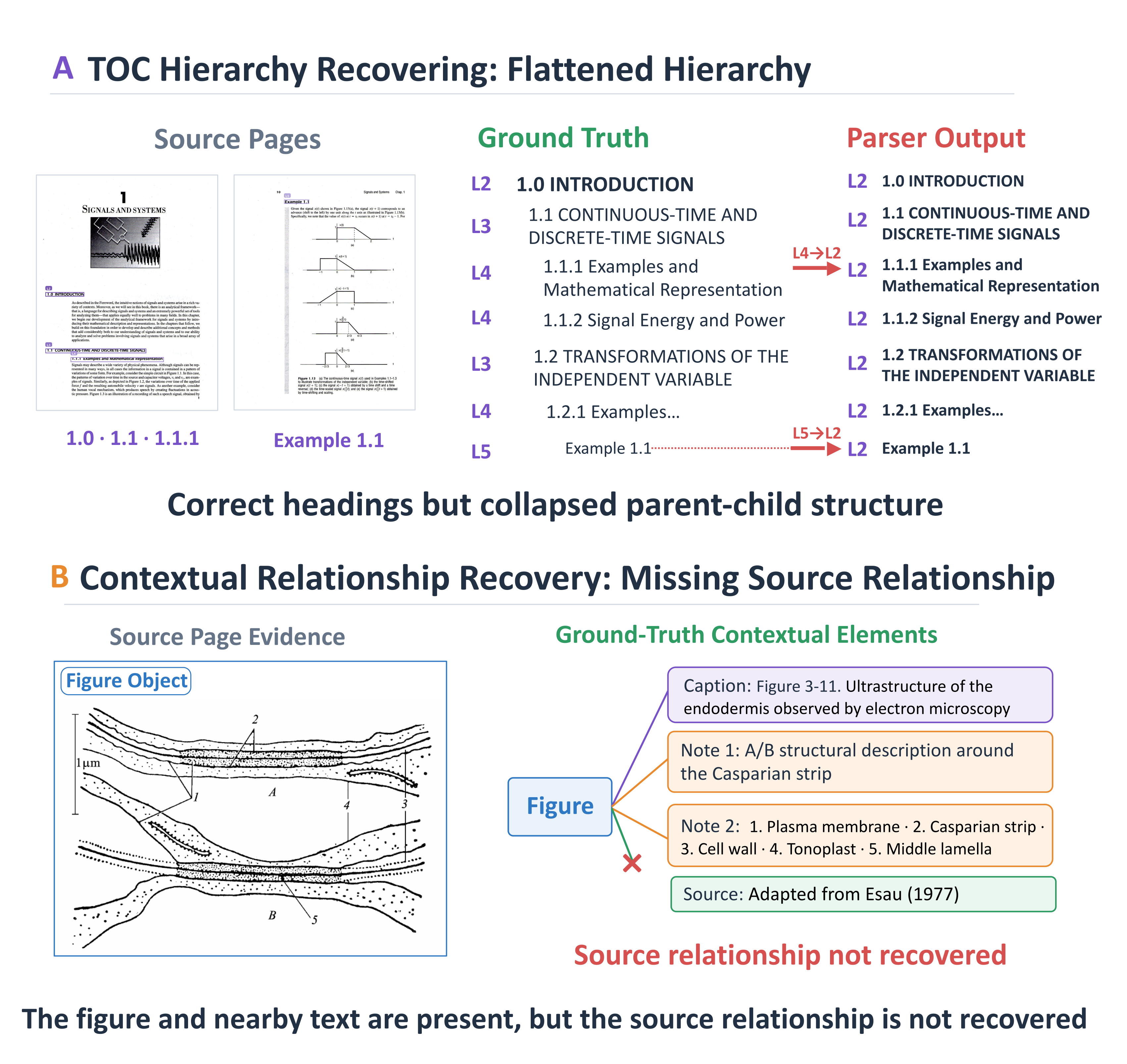}
    \caption{Representative challenges in document-level structure recovery. (A) A parser preserves heading text but collapses the parent--child structure of the TOC hierarchy. (B) A source relationship is not recovered despite the presence of the figure and nearby contextual text.}
    \label{fig:utility-challenges}
\end{figure}

\begin{figure*}[t]
    \centering
    \includegraphics[width=\textwidth]{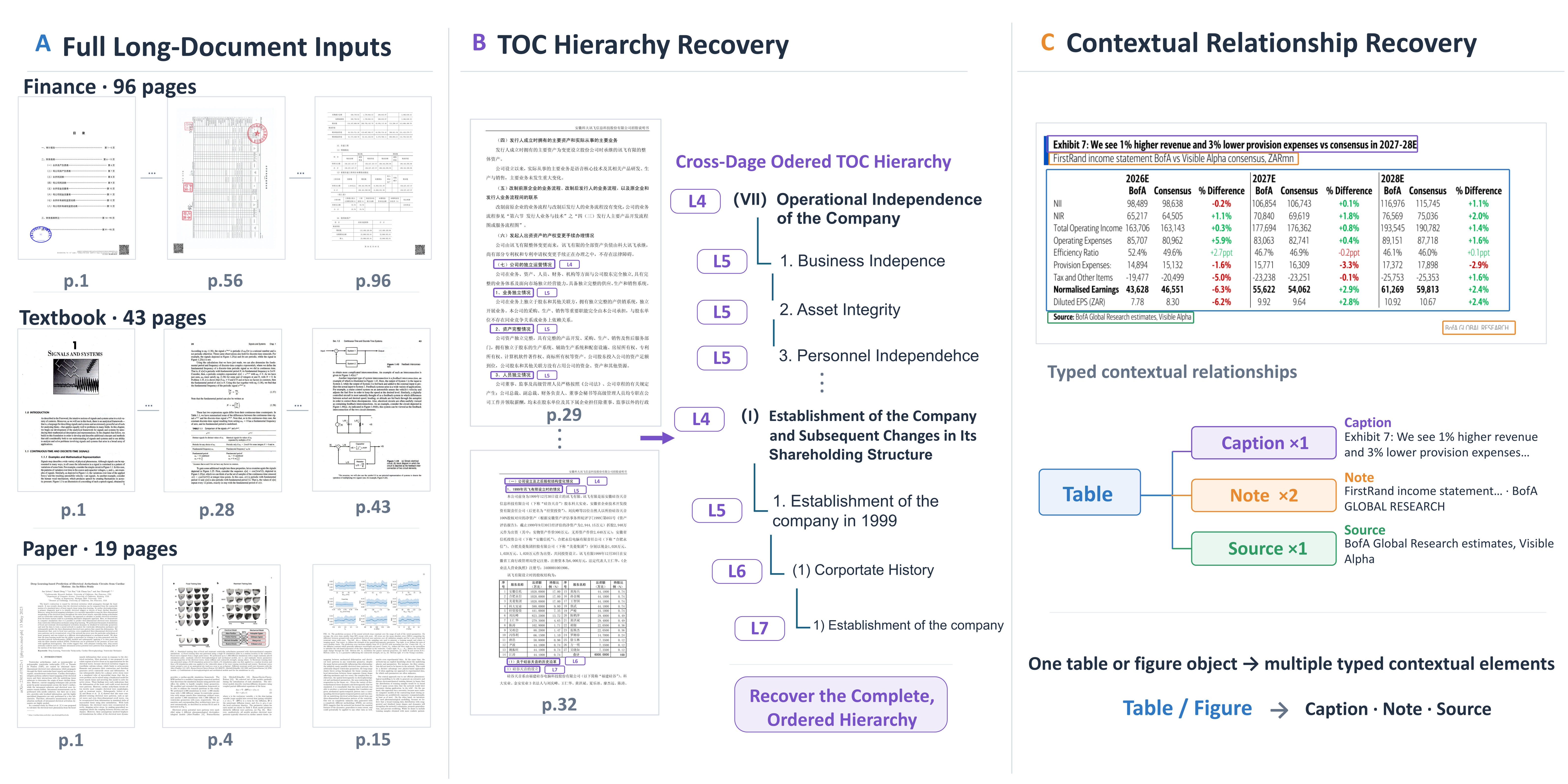}
    \caption{Overview of \textsc{LongDocBench}. (A) The benchmark contains complete long-document inputs from the finance, textbook, and paper domains. (B) \emph{Table-of-Contents Hierarchy Recovery} reconstructs complete, ordered TOC hierarchies across pages. (C) \emph{Contextual Relationship Recovery} identifies typed, potentially one-to-many links from tables and figures to their captions, notes, and sources.}
    \label{fig:longdocbench-overview}
\end{figure*}

\begin{figure*}[t]
    \centering
    \includegraphics[width=\textwidth]{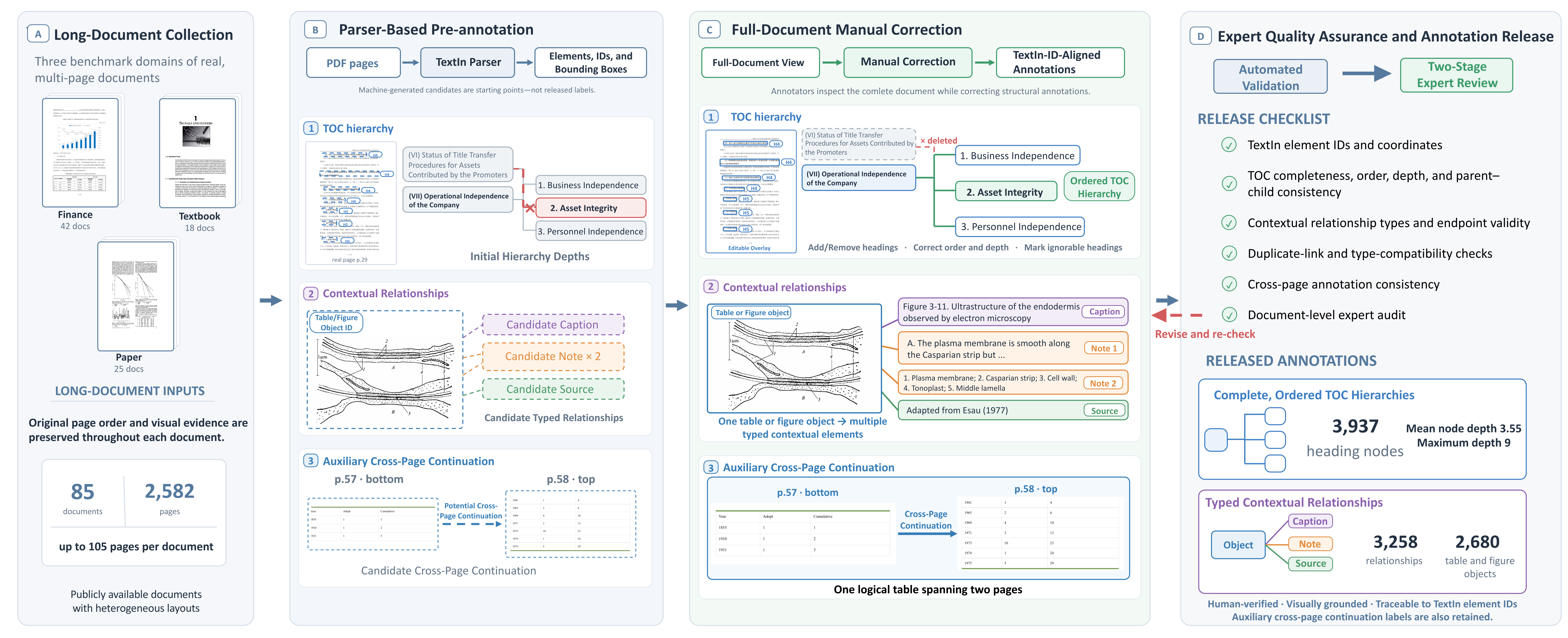}
    \caption{Construction and annotation pipeline of \textsc{LongDocBench}. TextIn generates candidate TOC hierarchies, typed contextual relationships, and auxiliary cross-page continuations. These candidates undergo full-document manual correction, automated validation, and two-stage expert review, yielding human-verified annotations that are visually grounded and traceable to TextIn element IDs.}
    \label{fig:construction-pipeline}
\end{figure*}

\section{Introduction}

Document parsing aims to convert visual documents into machine-readable representations that support information extraction, retrieval-augmented generation, and domain-specific analysis~\citep{xu2020layoutlm,lewis2020rag,mathew2021docvqa}. Advances in layout analysis, optical character recognition, and multimodal models have enabled increasingly reliable recognition of text, headings, formulas, tables, and figures on individual pages, as well as recovery of reading order and local layout structure~\citep{huang2022layoutlmv3,wang2021layoutreader,blecher2023nougat,wang2024mineru,ouyang2024omnidocbench,li2024readoc,zhou2026mpdocbench}. Long documents, however, require capabilities beyond page-level recognition. First, their content is often organized across tens or hundreds of pages through deeply nested table-of-contents (TOC) hierarchies, as commonly seen in financial reports and textbooks~\citep{bentabet2020fintoc,ma2023hrdoc,zhang2024pdftotree}. An incorrect parent--child assignment changes not only a heading's position in the hierarchy but also the sectional scope of the content governed by that heading. Second, tables and figures may be linked to captions, notes, and sources that are spatially distant, and a single object may be associated with multiple contextual elements~\citep{smock2022pubtables,xu2026minerupopo}. 
Therefore, Long-document parsing must consequently recover not only page-level elements, but also document-level TOC hierarchies and typed object-level contextual relationships from the document as a whole (Figure~\ref{fig:utility-challenges}).

Recent studies show that preserving multilevel document structure and recovering title hierarchies, cross-page continuity, and image--text associations improve downstream retrieval, generation, and question answering~\citep{sarthi2024raptor,buchmann2024documentstructure,saadfalcon2024pdftriage,lu2025hichunk,xu2026minerupopo}. Existing benchmarks, however, provide only partial diagnosis of such document organization: general parsing benchmarks emphasize page-level recognition, reading order, and formula or table structure, often through composite scores~\citep{zhong2019publaynet,zhong2020pubtabnet,pfitzmann2022doclaynet,wang2021layoutreader,ouyang2024omnidocbench}; multi-page benchmarks treat heading hierarchy and cross-page merging as dimensions within broader protocols~\citep{li2024readoc,zhou2026mpdocbench}; and specialized post-processing evaluations cover title-hierarchy reconstruction and image--text association without distinguishing caption, note, and source relationships~\citep{xu2026minerupopo}. Taken together, these limitations underscore the need for a unified, human-verified evaluation setting that directly assesses complete, ordered TOC hierarchies and provides type-specific evaluation of contextual relationships for tables and figures in real long documents. 

To enable more direct, fine-grained, and diagnostic evaluation of document-level structure recovery, we carefully design \textsc{LongDocBench} around two complementary tasks (Figure~\ref{fig:longdocbench-overview}): \emph{Table-of-Contents Hierarchy Recovery} and \emph{Contextual Relationship Recovery}. \textsc{LongDocBench} contains 85 real-world financial reports, textbooks, and academic papers spanning 2,582 pages, with up to 105 pages per document. It provides human-verified annotations for 3,937 heading nodes, with a mean node depth of 3.55 and a maximum depth of 9, as well as 3,258 typed contextual relationships involving 2,680 table and figure objects.

To examine whether these structures benefit long-document understanding, we conduct long-document question-answering experiments. The results show that both TOC hierarchies and contextual relationships improve question-answering accuracy, while their joint use yields further gains, indicating that they provide complementary organizational information. Although representative document parsers achieve strong and tightly clustered performance on conventional page-level parsing metrics, they remain substantially weaker at document-level structure recovery, reaching at best 0.55 Macro TEDS for TOC hierarchy recovery and a 0.63 Macro score for contextual relationship recovery. These results show that strong page-level parsing performance does not yet translate into reliable recovery of document-level TOC hierarchies and contextual relationships.

The key contributions of this work are threefold:
\begin{itemize}

\item We introduce and publicly release \textsc{LongDocBench}, a benchmark and unified evaluation protocol for two document-level structure-recovery tasks: \emph{Table-of-Contents Hierarchy Recovery} and \emph{Contextual Relationship Recovery}. It provides complete TOC hierarchies and type-specific contextual relationships for tables and figures, supporting reproducible and fine-grained evaluation.

\item We evaluate the downstream utility of these structures through long-document question-answering experiments. Both TOC hierarchies and contextual relationships improve question-answering accuracy, while their joint use yields further gains, indicating that they provide complementary organizational information.

\item We evaluate representative document parsers on both structure-recovery tasks and find that, despite strong page-level parsing performance, they remain limited in recovering complete TOC hierarchies and typed contextual relationships, indicating that both tasks remain challenging.

\end{itemize}

\section{LongDocBench}

To evaluate document-level structure recovery in long-document parsing, we introduce \textsc{LongDocBench}, a benchmark for two tasks: \emph{Table-of-Contents Hierarchy Recovery} and \emph{Contextual Relationship Recovery}. The former reconstructs headings distributed across pages into complete, ordered TOC hierarchies, while the latter identifies typed links between tables or figures and their captions, notes, and sources. To ensure diversity and practical relevance, \textsc{LongDocBench} comprises 85 real-world financial reports, textbooks, and academic papers spanning 2,582 pages, covering complex hierarchical organization, heterogeneous layouts, and rich contextual relationships for tables and figures.

\subsection{Data Collection}

We manually collect publicly accessible long-document PDFs from five source families: Chinese financial materials, international research reports, U.S. public company filings, textbooks, and academic papers. The collection includes public disclosures obtained through CNINFO and SEC EDGAR, together with scholarly articles identified through arXiv and DOI metadata~\citep{cninfo2026,sec2026edgar,arxiv2026,crossref2026}. The first three are grouped into the \emph{finance} domain, while textbooks and academic papers constitute the \emph{textbook} and \emph{paper} domains, respectively. Financial documents include annual and audit reports, prospectuses, debt-issuance announcements, regulatory filings, and institutional research reports, often featuring irregular layouts, deeply nested headings, and rich table/figure contextual relationships. Textbooks provide extended chapter structures across disciplines, whereas academic papers are generally shorter and contain regular sections with diverse formulas, tables, and figures.
Each document retains its original page order to preserve document-level organization and cross-page dependencies. The collection contains 42 financial documents, 18 textbooks, and 25 academic papers, totaling 85 documents and 2,582 pages, with up to 105 pages per document. It spans substantial variation in document length, layout, TOC depth, and table/figure contextual relationships.

\subsection{Data Annotation}

We construct the annotations through three stages (Figure~\ref{fig:construction-pipeline}): \emph{Parser-Based Pre-annotation}, \emph{Full-Document Manual Correction}, and \emph{Expert Quality Assurance and Annotation Release}. Manual correction is performed by trained annotators with backgrounds in document intelligence and visual document analysis, followed by automated validation and two-stage expert review by senior researchers with experience in document parsing, OCR, and structured document analysis.

\paragraph{Parser-Based Pre-annotation.}
TextIn extracts page-level elements, including text, element types, page indices, IDs, and bounding boxes. Detected headings are organized into initial TOC hierarchies, while table and figure objects are assigned candidate caption, note, and source relationships. Potential cross-page continuations are also identified. All machine-generated candidates serve only as starting points for manual correction.

\begin{figure}[!t]
    \centering
    \includegraphics[width=\columnwidth]{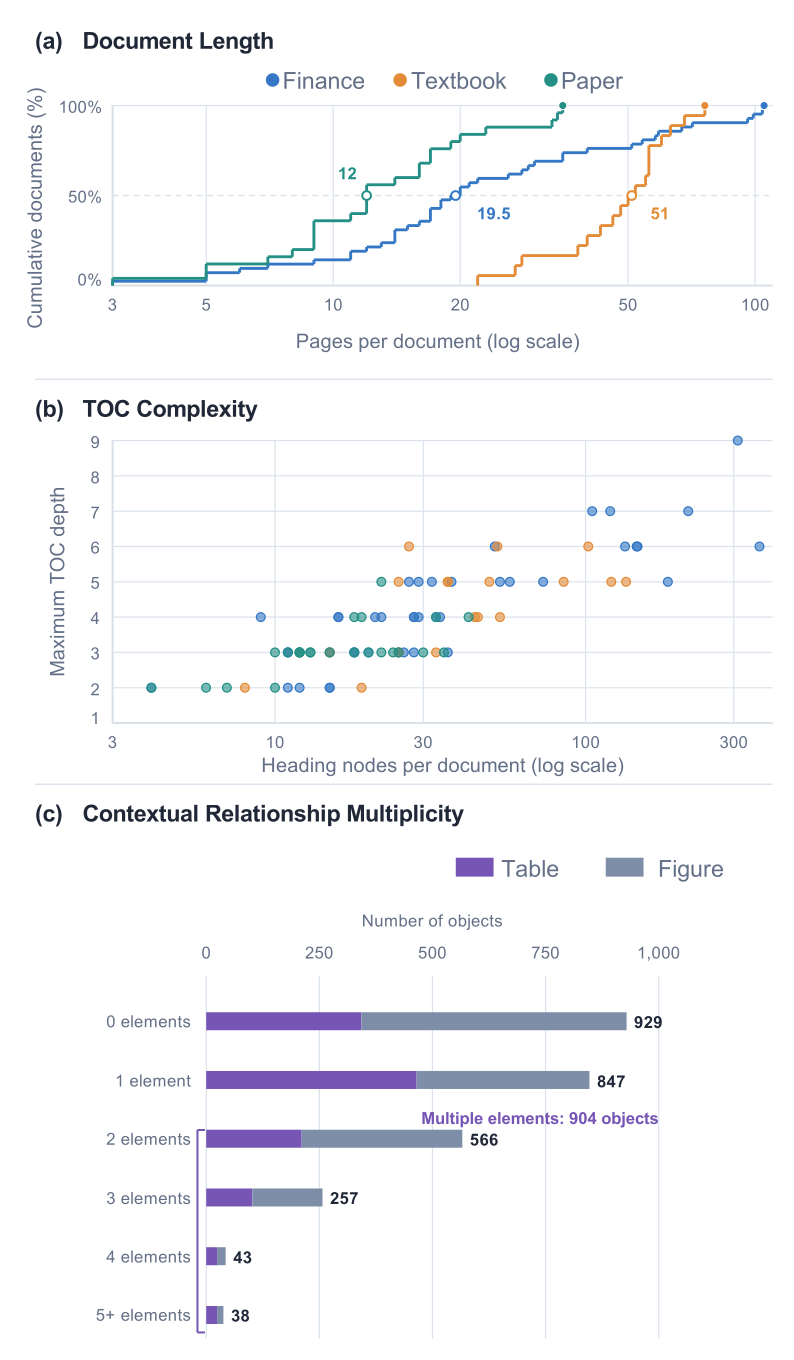}
    \caption{Dataset statistics of \textsc{LongDocBench}. (a) Document-length distributions; circles mark domain medians. (b) Per-document TOC complexity. (c) Number of contextual elements linked to each table or figure object.}
    \label{fig:dataset-statistics}
\end{figure}

\paragraph{Full-Document Manual Correction.}
Annotators inspect each complete document in a browser-based interface. They add or remove headings, correct heading text, order, depth, and coordinates, and mark ambiguous headings as \emph{ignorable}. They also revise contextual links, classify them as captions, notes, or sources, and verify one-to-many, shared, and cross-page relationships.

\paragraph{Expert Quality Assurance and Annotation Release.}
Automated validation checks element IDs and coordinates, hierarchy consistency, relationship endpoints, duplicate links, type compatibility, and cross-page consistency. Senior researchers then conduct two-stage expert review, including a final document-level audit. Failed documents are revised and re-checked before release. The final annotations contain 3,937 human-verified heading nodes and 3,258 typed contextual relationships annotated across 2,680 table and figure objects, together with auxiliary cross-page continuation labels.

\subsection{Dataset Statistics}

\paragraph{Document Distribution.}
As shown in Table~\ref{tab:dataset-stats-en} and Figure~\ref{fig:dataset-statistics}(a), \textsc{LongDocBench} contains 85 documents spanning 2,582 pages, with an average of 30.38 pages per document. The finance domain is the largest, comprising 42 documents and 1,331 pages, and contains the longest document at 105 pages. Although the textbook domain contains only 18 documents, it has the greatest average length at 49.11 pages. Documents in the paper domain are comparatively compact, averaging 14.68 pages across 25 documents. The collection therefore spans substantial variation in document length across domains.

\begin{table}[t]
    \centering
    \small
    \setlength{\tabcolsep}{3pt}
    \begin{tabular}{lcccc}
        \toprule
        Domain & Docs & Pages & Avg. pages/doc. & Max pages \\
        \midrule
        Finance & 42 & 1,331 & 31.69 & 105 \\
        Textbook & 18 & 884 & 49.11 & 76 \\
        Paper & 25 & 367 & 14.68 & 35 \\
        \midrule
        All & \textbf{85} & \textbf{2,582} & \textbf{30.38} & \textbf{105} \\
        \bottomrule
    \end{tabular}
    \caption{Document statistics by domain.}
    \label{tab:dataset-stats-en}
\end{table}

\paragraph{TOC Hierarchies.}
Table~\ref{tab:toc-stats-en} summarizes the 3,937 annotated heading nodes. Depth is one-indexed, and mean depth is computed over all heading nodes within each domain. The finance domain contains 2,582 nodes with a mean depth of 3.81 and a maximum depth of 9, forming the largest and deepest TOC hierarchies in the benchmark. The textbook domain contains 910 nodes with a mean depth of 3.31 and a maximum depth of 6, while the paper domain contains 445 nodes with a mean depth of 2.49 and a maximum depth of 5. These distributions support evaluation across both shallow section structures and deeply nested hierarchies.

\begin{table}[t]
    \centering
    \small
    \setlength{\tabcolsep}{4pt}
    \begin{tabular}{lccc}
        \toprule
        Domain & Heading nodes & Mean node depth & Max depth \\
        \midrule
        Finance & 2,582 & 3.81 & 9 \\
        Textbook & 910 & 3.31 & 6 \\
        Paper & 445 & 2.49 & 5 \\
        \midrule
        All & \textbf{3,937} & \textbf{3.55} & \textbf{9} \\
        \bottomrule
    \end{tabular}
    \caption{TOC hierarchy statistics by domain.}
    \label{tab:toc-stats-en}
\end{table}

\paragraph{Contextual Relationships.}
As reported in Table~\ref{tab:relation-stats-en}, the benchmark contains 1,169 tables and 1,511 figures, yielding 2,680 annotated objects. These objects are connected by 3,258 typed contextual relationships, including 1,703 caption, 961 note, and 594 source relationships. Among them, 904 objects are linked to multiple contextual elements, directly demonstrating the prevalence of one-to-many associations. Tables have more note relationships than figures (587 versus 374), whereas figures have more source relationships (388 versus 206), reflecting different contextual patterns across object types. Together, these statistics provide diverse challenges for TOC hierarchy recovery and contextual relationship recovery.

\begin{table}[t]
    \centering
    \small
    \setlength{\tabcolsep}{4pt}
    \begin{tabular}{lccc}
        \toprule
        Category & Table & Figure & Total \\
        \midrule
        Objects & 1,169 & 1,511 & 2,680 \\
        \midrule
        Caption relationships & 713 & 990 & 1,703 \\
        Note relationships & 587 & 374 & 961 \\
        Source relationships & 206 & 388 & 594 \\
        \midrule
        All relationships & \textbf{1,506} & \textbf{1,752} & \textbf{3,258} \\
        \bottomrule
    \end{tabular}
    \caption{Object and contextual-relationship statistics.}
    \label{tab:relation-stats-en}
\end{table}

\section{Experiments}

\subsection{Evaluation Metrics}

\paragraph{TOC Hierarchy Recovery.}
We evaluate complete, ordered heading hierarchies using Tree Edit Distance Similarity (TEDS), which jointly reflects errors in heading content, order, and parent--child structure~\citep{zhang1989simple,zhong2020pubtabnet}. We report \emph{Macro TEDS}, averaged equally across documents, and \emph{Weighted TEDS}, weighted by document page count, under both with- and without-ignorable settings.

\paragraph{Contextual Relationship Recovery.}
Predicted tables and figures are matched to same-type ground-truth objects on the same page based on bounding-box IoU. For each matched object, caption, note, and source relationships are evaluated using normalized text edit similarity~\citep{levenshtein1966binary}. We report the score for each relationship type and their unweighted mean as the final \emph{Macro score}.

Detailed normalization, matching, aggregation, and scoring procedures are provided in the appendix.

\subsection{Experimental Setup}

We evaluate TOC hierarchy recovery on all 85 documents using TextIn, MinerU2.5-Pro, GLM-OCR, PaddleOCR-VL-1.5, and PaddleOCR-VL-1.6, converting their native heading outputs into ordered trees through a unified construction procedure~\citep{wang2024mineru,duan2026glmocr,cui2026paddleocrvl15,zhang2026paddleocrvl16}. For contextual relationship recovery, we fix TextIn parsing and benchmark GPT-5.6-Sol, MiniMax-M2.5, GLM-5.2, Kimi K2.6, and Qwen3.5 variants on 2,680 benchmark-localized table and figure targets, using the parsed text and layout context to predict caption, note, and source relationships~\citep{minimax2026m2,glm5team2026glm5,kimi2025k2,yang2025qwen3}. Using the OmniDocBench v1.6 protocol, we evaluate the page-level parsing performance of LingDT-VL-OCR-4B, MinerU2.5-Pro, PaddleOCR-VL-1.5, PaddleOCR-VL-1.6, and GLM-OCR on all 2,582 pages of \textsc{LongDocBench}~\citep{ouyang2024omnidocbench,qian2026lingdt}. To assess reasoning utility, we compare a structure-agnostic fixed-chunk BM25 baseline with structure-aware question answering~\citep{lewis2020rag,robertson2009bm25}. The recovered-structure conditions use TOC hierarchies produced by TextIn and contextual relationships produced by Qwen3.5-397B with thinking, while the verified conditions use human-verified \textsc{LongDocBench} structures. All conditions use the same parsed content, questions, answer prompt, and GLM-5.2 answer model. Detailed experiment settings are provided in the appendix.

\subsection{TOC Hierarchy Recovery}

\begin{figure*}[t]
    \centering
    \includegraphics[width=\textwidth]{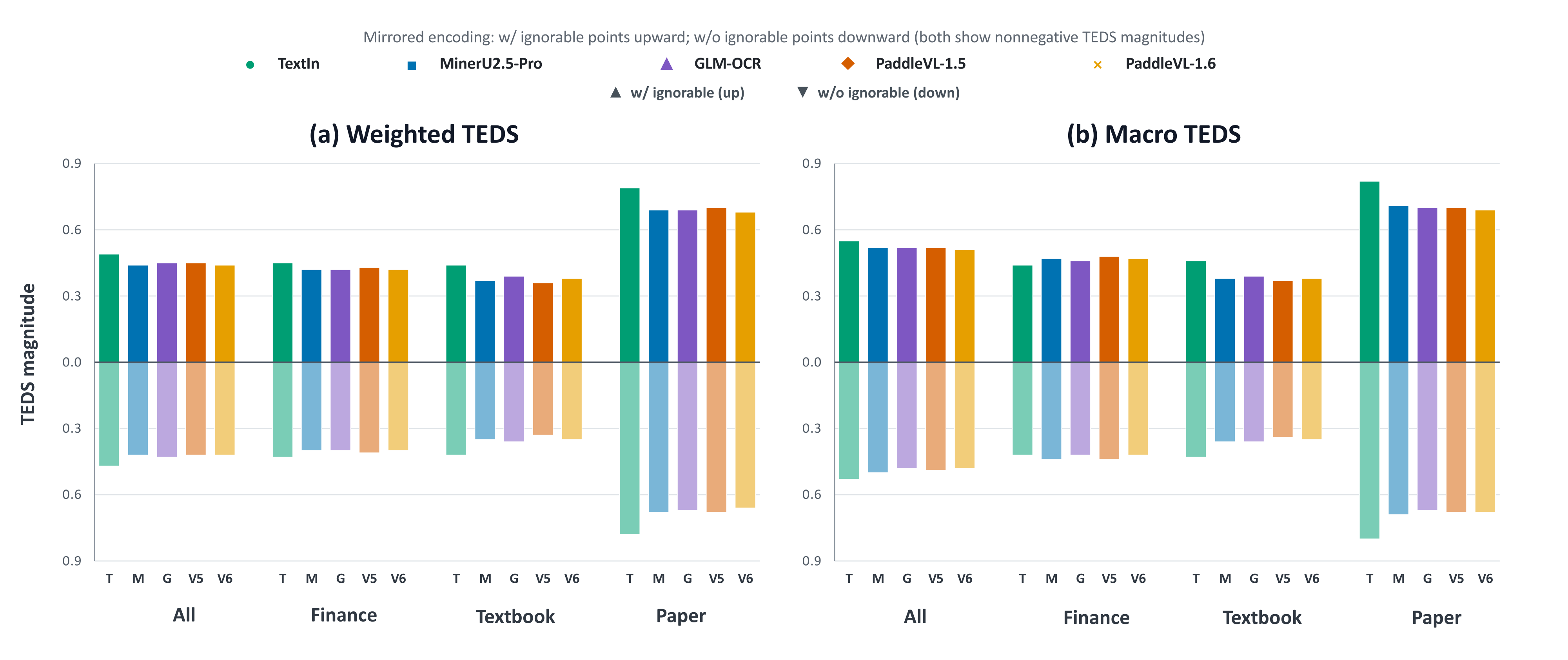}
    \caption{TOC hierarchy recovery by domain. Panels report (a) Weighted TEDS and (b) Macro TEDS. For each model and domain, the with-ignorable score is plotted upward and the without-ignorable score is mirrored downward; downward bars therefore encode nonnegative magnitudes rather than negative TEDS. Higher magnitudes indicate better recovery.}
    \label{fig:toc-results}
\end{figure*}

We first evaluate whether document parsers that perform well at page-level element recognition can reconstruct complete, ordered TOC hierarchies across pages. Table~\ref{tab:toc-results-en} reports the overall results, while Figure~\ref{fig:toc-results} compares performance across document domains. TextIn performs best, achieving 0.49 Weighted TEDS and 0.55 Macro TEDS under the with-ignorable setting. The remaining systems are closely clustered, with Weighted TEDS of 0.44--0.45 and Macro TEDS of 0.51--0.52. These scores indicate that recovering document-level heading organization remains difficult even for strong document parsers.

Performance varies markedly across domains. Macro TEDS reaches 0.69--0.82 on papers but only 0.44--0.48 on financial documents and 0.37--0.46 on textbooks. This pattern is consistent with the dataset statistics: papers are generally shorter and have shallower section structures, whereas financial documents and textbooks contain longer and more deeply nested hierarchies. Weighted TEDS is consistently lower than Macro TEDS, further indicating that recovery quality deteriorates as document length increases.

Accounting for ignorable headings improves Macro TEDS by only 0.02--0.04 and Weighted TEDS by 0.02--0.03, without materially changing system rankings or domain-level trends. The remaining errors therefore arise primarily from failures to recover heading order, depth, and parent--child structure rather than from optional or ambiguous headings.

\begin{table}[t]
    \centering
    \small
    \setlength{\tabcolsep}{3pt}
        \begin{tabular}{lcccc}
            \toprule
            Model
            & \multicolumn{2}{c}{Weighted TEDS}
            & \multicolumn{2}{c}{Macro TEDS} \\
            \cmidrule(lr){2-3}
            \cmidrule(lr){4-5}
            & w/ ign. & w/o ign.
            & w/ ign. & w/o ign. \\
            \midrule
            TextIn           & 0.49 & 0.47 & 0.55 & 0.53 \\
            MinerU2.5-Pro    & 0.44 & 0.42 & 0.52 & 0.50 \\
            GLM-OCR          & 0.45 & 0.43 & 0.52 & 0.48 \\
            PaddleOCR-VL-1.5 & 0.45 & 0.42 & 0.52 & 0.49 \\
            PaddleOCR-VL-1.6 & 0.44 & 0.42 & 0.51 & 0.48 \\
            \bottomrule
        \end{tabular}
    \caption{TOC hierarchy recovery on all 85 documents. We report page-count-weighted and document-macro TEDS under the with- and without-ignorable settings; higher is better.}
    \label{tab:toc-results-en}
\end{table}

\subsection{Contextual Relationship Recovery}

We next isolate contextual relationship recovery from object detection by providing all models with the same TextIn parsing outputs and benchmark-localized table and figure objects. As shown in Table~\ref{tab:relation-results-en}, GPT-5.6-Sol achieves the highest Macro score of 0.63, followed by Qwen3.5-397B with thinking at 0.60. No model dominates all relation types: GPT-5.6-Sol performs best on captions and notes, whereas Qwen3.5-397B without thinking achieves the highest source score. The limited best Macro score shows that reliable contextual relationship recovery remains difficult even when object locations and page-level parsing are fixed.

The difficulty also differs substantially by relationship type. Source relationships obtain the highest scores, ranging from 0.61 to 0.85, followed by captions at 0.53--0.65. Notes are consistently the most challenging, with scores of only 0.16--0.42. Enabling thinking for Qwen3.5-397B raises its Macro score only marginally, from 0.59 to 0.60, and does not resolve its weakness on notes. These results identify note recovery as the principal bottleneck, reflecting the greater variability and weaker spatial regularity of note relationships.

\begin{table}[t]
\centering
\small
\setlength{\tabcolsep}{2.5pt}
    \begin{tabular}{lcccc}
        \toprule
        Recovery model & Caption & Note & Source & Macro \\
        \midrule
        GPT-5.6-Sol (xhigh)
        & \textbf{0.65} & \textbf{0.42} & 0.81 & \textbf{0.63} \\
        MiniMax-M2.5 (think)
        & 0.53 & 0.17 & 0.61 & 0.44 \\
        GLM-5.2 (think)
        & 0.64 & 0.30 & 0.74 & 0.56 \\
        Kimi K2.6 (w/o think)
        & 0.62 & 0.22 & 0.82 & 0.55 \\
        Qwen3.5-397B (w/o think)
        & 0.61 & 0.30 & \textbf{0.85} & 0.59 \\
        Qwen3.5-397B (think)
        & 0.64 & 0.33 & 0.82 & 0.60 \\
        Qwen3.5-35B (w/o think)
        & 0.57 & 0.20 & 0.78 & 0.52 \\
        Qwen3.5-9B (w/o think)
        & 0.59 & 0.16 & 0.80 & 0.52 \\
        \bottomrule
    \end{tabular}
\caption{Contextual relationship recovery using common TextIn outputs and benchmark-localized objects. Columns score individual relationship types; Macro is their unweighted mean. Higher is better.}
\label{tab:relation-results-en}
\end{table}

\begin{table*}[t]
    \centering
    \small
    \setlength{\tabcolsep}{1mm}
    \begin{tabular}{@{}lrrrrrr@{}}
        \toprule
        Model & Text $\downarrow$ & Formula (CDM) $\uparrow$
        & Table (TEDS) $\uparrow$ & Table (TEDS-S) $\uparrow$
        & Reading Order $\downarrow$ & Overall $\uparrow$ \\
        \midrule
        LingDT-VL-OCR-4B
        & 0.02 & 95.37 & 91.75 & 93.90 & \textbf{0.10} & 95.07 \\
        MinerU2.5-Pro
        & \textbf{0.01} & \textbf{95.78} & \textbf{93.33}
        & \textbf{94.75} & 0.11 & \textbf{95.96} \\
        PaddleOCR-VL-1.5
        & 0.04 & 95.36 & 89.15 & 91.45 & 0.13 & 93.54 \\
        PaddleOCR-VL-1.6
        & 0.04 & 95.50 & 89.67 & 92.12 & 0.13 & 93.67 \\
        GLM-OCR
        & 0.03 & 95.71 & 88.76 & 91.25 & 0.14 & 93.86 \\
        \bottomrule
    \end{tabular}
    \caption{Page-level document parsing on all 2,582 pages under the OmniDocBench v1.6 protocol. Down arrows denote error metrics (lower is better), whereas up arrows denote similarity or accuracy metrics (higher is better).}
    \label{tab:omnidoc-results-en}
\end{table*}

\subsection{Document Parsing Results}

We further evaluate whether strong conventional parsing performance is accompanied by reliable document-level structure recovery. Table~\ref{tab:omnidoc-results-en} reports page-level parsing results under the OmniDocBench v1.6 protocol~\citep{ouyang2024omnidocbench,zhong2020pubtabnet}. All five systems achieve strong and closely clustered Overall scores of 93.54--95.96. MinerU2.5-Pro obtains the best Overall score and leads on text, formula, and table parsing, while LingDT-VL-OCR-4B achieves the lowest reading-order error. Formula scores vary little across systems, whereas table parsing exhibits somewhat larger differences.

These results contrast sharply with performance on the two document-level tasks. Despite Overall page-level scores above 93.5, the best systems reach only 0.55 Macro TEDS for TOC hierarchy recovery and 0.63 Macro for contextual relationship recovery. Strong recognition of page elements and local structure therefore does not imply reliable reconstruction of cross-page heading hierarchies or typed table/figure relationships. Document-level structure recovery remains a distinct capability that is not captured by conventional page-level parsing metrics.

\subsection{Impact on Long-Document Reasoning}

\begin{table}[!t]
    \centering
    \small
    \setlength{\tabcolsep}{3pt}
    \begin{tabular}{@{}lr@{}}
        \toprule
        Condition & Answer Acc. \\
        \midrule
        Fixed + BM25 & 32.98 \\
        \midrule
        TextIn TOC & 30.83 \\
        Verified TOC & 45.95 \\
        Qwen3.5 Relations & 31.90 \\
        Verified Relations & 47.62 \\
        TextIn TOC + Qwen3.5 Relations & 37.86 \\
        Verified TOC + Relations & \textbf{54.29} \\
        \bottomrule
    \end{tabular}
    \caption{Test-set Combined Answer Accuracy (\%). Fixed + BM25 is the structure-agnostic chunk baseline. TextIn TOC and Qwen3.5 Relations use recovered structures, whereas Verified conditions use human-verified structures. All conditions share the same parsed content and answer model. The best result is shown in bold.}
    \label{tab:qa-effects-en}
\end{table}

We compare the structure-agnostic Fixed + BM25 baseline with conditions using recovered or human-verified TOC hierarchies and contextual relationships. All conditions share the same parsed document content and answer model. Table~\ref{tab:qa-effects-en} reports the test-set results.

Human-verified structures provide substantial gains over the Fixed + BM25 baseline of 32.98\%. Verified TOC and Verified Relations achieve 45.95\% and 47.62\%, improving accuracy by 12.97 and 14.64 percentage points, respectively. Combining both structures yields the best result of 54.29\%, a gain of 21.31 points over the baseline and additional gains of 8.34 and 6.67 points over Verified TOC and Verified Relations alone. These results indicate that TOC hierarchies and contextual relationships contribute complementary organizational information to long-document reasoning.

Automatically recovered structures realize only part of this potential. TextIn TOC and Qwen3.5 Relations obtain 30.83\% and 31.90\%, neither surpassing the fixed-chunk baseline, while their combination reaches 37.86\%, exceeding the baseline by 4.88 points. The recovered conditions remain 15.12, 15.72, and 16.43 points below their verified TOC, relation, and combined counterparts, respectively. Thus, the verified results establish the utility of the two structures, whereas the recovered-to-verified gaps show that current recovery quality remains insufficient to realize their full downstream benefit.

\section{Conclusion}

In this paper, we introduced \textsc{LongDocBench}, a human-verified benchmark for two document-level structure-recovery tasks: \emph{Table-of-Contents Hierarchy Recovery} and \emph{Contextual Relationship Recovery}. It contains 85 real-world documents spanning 2,582 pages, with 3,937 heading nodes and 3,258 typed contextual relationships annotated across 2,680 table and figure objects. We further demonstrate the downstream value of these structures: human-verified TOC hierarchies and contextual relationships jointly improve test-set question-answering accuracy from 32.98\% with fixed-chunk BM25 retrieval to 54.29\%. Finally, representative parsers achieve only 0.55 Macro TEDS and a 0.63 Macro score on the two recovery tasks despite strong page-level performance, while automatically recovered structures provide much smaller reasoning gains. Although limited in scale and coverage, \textsc{LongDocBench} establishes document-level organization recovery as a useful, insufficiently addressed capability.

\bibliography{longdocbench_references}

@inproceedings{ouyang2024omnidocbench,
  title        = {{OmniDocBench}: Benchmarking Diverse {PDF} Document Parsing with Comprehensive Annotations},
  author       = {Ouyang, Linke and Qu, Yuan and Zhou, Hongbin and Zhu, Jiawei and Zhang, Rui and Lin, Qunshu and Wang, Bin and Zhao, Zhiyuan and Jiang, Man and Zhao, Xiaomeng and Shi, Jin and Wu, Fan and Chu, Pei and Liu, Minghao and Li, Zhenxiang and Xu, Chao and Zhang, Bo and Shi, Botian and Tu, Zhongying and He, Conghui},
  booktitle    = {Proceedings of the IEEE/CVF Conference on Computer Vision and Pattern Recognition},
  pages        = {24838--24848},
  year         = {2025}
}

@article{zhou2026mpdocbench,
  title        = {{MPDocBench-Parse}: Benchmarking Practical Multi-page Document Parsing},
  author       = {Zhou, Bangbang and Xing, Hangdi and Chen, Yifan and Xu, Jianjun and Zheng, Qi and Gao, Feiyu and Yang, Zhibo and Bai, Shuai and Yan, Ming and Ye, Jieping and Xie, Hongtao},
  journal      = {arXiv preprint arXiv:2605.22100},
  year         = {2026}
}

@article{lu2025hichunk,
  title        = {{HiChunk}: Evaluating and Enhancing Retrieval-Augmented Generation with Hierarchical Chunking},
  author       = {Lu, Wensheng and Chen, Keyu and Qiao, Ruizhi and Sun, Xing},
  journal      = {arXiv preprint arXiv:2509.11552},
  year         = {2025}
}

@article{wang2024mineru,
  title        = {{MinerU}: An Open-Source Solution for Precise Document Content Extraction},
  author       = {Wang, Bin and Xu, Chao and Zhao, Xiaomeng and Ouyang, Linke and Wu, Fan and Zhao, Zhiyuan and Xu, Rui and Liu, Kaiwen and Qu, Yuan and Shang, Fukai and Zhang, Bo and Wei, Liqun and Sui, Zhihao and Li, Wei and Shi, Botian and Qiao, Yu and Lin, Dahua and He, Conghui},
  journal      = {arXiv preprint arXiv:2409.18839},
  year         = {2024}
}

@article{yang2025qwen3,
  title        = {{Qwen3} Technical Report},
  author       = {Yang, An and Li, Anfeng and Yang, Baosong and Zhang, Beichen and Hui, Binyuan and Zheng, Bo and Yu, Bowen and Gao, Chang and Huang, Chengen and Lv, Chenxu and others},
  journal      = {arXiv preprint arXiv:2505.09388},
  year         = {2025}
}

@inproceedings{zhong2019publaynet,
  title        = {{PubLayNet}: Largest Dataset Ever for Document Layout Analysis},
  author       = {Zhong, Xu and Tang, Jianbin and Jimeno Yepes, Antonio},
  booktitle    = {2019 International Conference on Document Analysis and Recognition (ICDAR)},
  pages        = {1015--1022},
  publisher    = {IEEE},
  year         = {2019},
  doi          = {10.1109/ICDAR.2019.00166}
}

@article{pfitzmann2022doclaynet,
  title        = {{DocLayNet}: A Large Human-Annotated Dataset for Document-Layout Analysis},
  author       = {Pfitzmann, Birgit and Auer, Christoph and Dolfi, Michele and Nassar, Ahmed S. and Staar, Peter W. J.},
  journal      = {arXiv preprint arXiv:2206.01062},
  year         = {2022}
}

@inproceedings{zhong2020pubtabnet,
  title        = {Image-Based Table Recognition: Data, Model, and Evaluation},
  author       = {Zhong, Xu and ShafieiBavani, Elaheh and Jimeno Yepes, Antonio},
  booktitle    = {Computer Vision -- ECCV 2020},
  pages        = {564--580},
  publisher    = {Springer},
  year         = {2020},
  doi          = {10.1007/978-3-030-58589-1_34}
}

@inproceedings{smock2022pubtables,
  title        = {{PubTables-1M}: Towards Comprehensive Table Extraction from Unstructured Documents},
  author       = {Smock, Brandon and Pesala, Rohith and Abraham, Robin},
  booktitle    = {Proceedings of the IEEE/CVF Conference on Computer Vision and Pattern Recognition},
  pages        = {4634--4642},
  year         = {2022}
}

@inproceedings{xu2020layoutlm,
  title        = {{LayoutLM}: Pre-Training of Text and Layout for Document Image Understanding},
  author       = {Xu, Yiheng and Li, Minghao and Cui, Lei and Huang, Shaohan and Wei, Furu and Zhou, Ming},
  booktitle    = {Proceedings of the 26th ACM SIGKDD International Conference on Knowledge Discovery and Data Mining},
  pages        = {1192--1200},
  publisher    = {ACM},
  year         = {2020},
  doi          = {10.1145/3394486.3403172}
}

@inproceedings{huang2022layoutlmv3,
  title        = {{LayoutLMv3}: Pre-Training for Document {AI} with Unified Text and Image Masking},
  author       = {Huang, Yupan and Lv, Tengchao and Cui, Lei and Lu, Yutong and Wei, Furu},
  booktitle    = {Proceedings of the 30th ACM International Conference on Multimedia},
  pages        = {4083--4091},
  publisher    = {ACM},
  year         = {2022},
  doi          = {10.1145/3503161.3548112}
}

@inproceedings{blecher2023nougat,
  title        = {{Nougat}: Neural Optical Understanding for Academic Documents},
  author       = {Blecher, Lukas and Cucurull, Guillem and Scialom, Thomas and Stojnic, Robert},
  booktitle    = {International Conference on Learning Representations},
  year         = {2024},
  url          = {https://openreview.net/forum?id=fUtxNAKpdV}
}

@inproceedings{lewis2020rag,
  title        = {Retrieval-Augmented Generation for Knowledge-Intensive {NLP} Tasks},
  author       = {Lewis, Patrick and Perez, Ethan and Piktus, Aleksandra and Petroni, Fabio and Karpukhin, Vladimir and Goyal, Naman and K{\"u}ttler, Heinrich and Lewis, Mike and Yih, Wen-tau and Rockt{\"a}schel, Tim and Riedel, Sebastian and Kiela, Douwe},
  booktitle    = {Advances in Neural Information Processing Systems},
  volume       = {33},
  pages        = {9459--9474},
  year         = {2020}
}

@inproceedings{sarthi2024raptor,
  title        = {{RAPTOR}: Recursive Abstractive Processing for Tree-Organized Retrieval},
  author       = {Sarthi, Parth and Abdullah, Salman and Tuli, Aditi and Khanna, Shubh and Goldie, Anna and Manning, Christopher D.},
  booktitle    = {International Conference on Learning Representations},
  year         = {2024},
  url          = {https://openreview.net/forum?id=GN921JHCRw}
}

@inproceedings{mathew2021docvqa,
  title        = {{DocVQA}: A Dataset for {VQA} on Document Images},
  author       = {Mathew, Minesh and Karatzas, Dimosthenis and Jawahar, C. V.},
  booktitle    = {Proceedings of the IEEE/CVF Winter Conference on Applications of Computer Vision},
  pages        = {2200--2209},
  year         = {2021}
}

@inproceedings{bentabet2020fintoc,
  title        = {The Financial Document Structure Extraction Shared Task: {FinTOC} 2020},
  author       = {Bentabet, Najah-Imane and Juge, R{\'e}mi and El Maarouf, Ismail and Mouilleron, Virginie and Valsamou-Stanislawski, Dialekti and El-Haj, Mahmoud},
  booktitle    = {Proceedings of the 1st Joint Workshop on Financial Narrative Processing and MultiLing Financial Summarisation},
  pages        = {13--22},
  publisher    = {COLING},
  year         = {2020},
  url          = {https://aclanthology.org/2020.fnp-1.2/}
}

@inproceedings{wang2021layoutreader,
  title        = {{LayoutReader}: Pre-Training of Text and Layout for Reading Order Detection},
  author       = {Wang, Zilong and Xu, Yiheng and Cui, Lei and Shang, Jingbo and Wei, Furu},
  booktitle    = {Proceedings of the 2021 Conference on Empirical Methods in Natural Language Processing},
  pages        = {4735--4744},
  publisher    = {Association for Computational Linguistics},
  year         = {2021},
  doi          = {10.18653/v1/2021.emnlp-main.389}
}

@inproceedings{ma2023hrdoc,
  title        = {{HRDoc}: Dataset and Baseline Method toward Hierarchical Reconstruction of Document Structures},
  author       = {Ma, Jiefeng and Du, Jun and Hu, Pengfei and Zhang, Zhenrong and Zhang, Jianshu and Zhu, Huihui and Liu, Cong},
  booktitle    = {Proceedings of the AAAI Conference on Artificial Intelligence},
  volume       = {37},
  pages        = {1870--1877},
  year         = {2023},
  doi          = {10.1609/aaai.v37i2.25277}
}

@inproceedings{zhang2024pdftotree,
  title        = {{PDF-to-Tree}: Parsing {PDF} Text Blocks into a Tree},
  author       = {Zhang, Yue and Zhang, Zhihao and Lai, Wenbin and Zhang, Chong and Gui, Tao and Zhang, Qi and Huang, Xuanjing},
  booktitle    = {Findings of the Association for Computational Linguistics: EMNLP 2024},
  pages        = {10704--10714},
  publisher    = {Association for Computational Linguistics},
  year         = {2024},
  doi          = {10.18653/v1/2024.findings-emnlp.628}
}

@article{qian2026lingdt,
  title        = {{LingDT-VL-OCR}: Structure-Aware Document-Level Parsing with Fine-Grained Visual Reference},
  author       = {Qian, Siyi and Bai, Xiongfei and Fu, Bingtao and Lu, Yichen and Zhang, Gaoyang and Yang, Xudong and Zhang, Peng},
  journal      = {arXiv preprint arXiv:2603.11044},
  year         = {2026}
}

@inproceedings{saadfalcon2024pdftriage,
  title        = {{PDFTriage}: Question Answering over Long, Structured Documents},
  author       = {Saad-Falcon, Jon and Barrow, Joe and Siu, Alexa and Nenkova, Ani and Yoon, Seunghyun and Rossi, Ryan A. and Dernoncourt, Franck},
  booktitle    = {Proceedings of the 2024 Conference on Empirical Methods in Natural Language Processing: Industry Track},
  pages        = {153--169},
  publisher    = {Association for Computational Linguistics},
  year         = {2024},
  doi          = {10.18653/v1/2024.emnlp-industry.13}
}

@inproceedings{buchmann2024documentstructure,
  title        = {Document Structure in Long Document Transformers},
  author       = {Buchmann, Jan and Eichler, Max and Bodensohn, Jan-Micha and Kuznetsov, Ilia and Gurevych, Iryna},
  booktitle    = {Proceedings of the 18th Conference of the European Chapter of the Association for Computational Linguistics (Volume 1: Long Papers)},
  pages        = {1056--1073},
  publisher    = {Association for Computational Linguistics},
  year         = {2024},
  doi          = {10.18653/v1/2024.eacl-long.64}
}

@article{li2024readoc,
  title        = {{READoc}: A Unified Benchmark for Realistic Document Structured Extraction},
  author       = {Li, Zichao and Abulaiti, Aizier and Lu, Yaojie and Chen, Xuanang and Zheng, Jia and Lin, Hongyu and Han, Xianpei and Sun, Le},
  journal      = {arXiv preprint arXiv:2409.05137},
  year         = {2024}
}

@article{xu2026minerupopo,
  title        = {{MinerU-Popo}: Universal Post-Processing Model for Structured Document Parsing},
  author       = {Xu, Bangrui and Miao, Ziyang and Zhou, Xuanhe and Lin, Yiming and Tang, Zirui and Zhao, Xiaomeng and Wu, Fan and Tan, Cheng and Wang, Bin and He, Conghui},
  journal      = {arXiv preprint arXiv:2605.24973},
  year         = {2026}
}

@article{zhang1989simple,
  title        = {Simple Fast Algorithms for the Editing Distance between Trees and Related Problems},
  author       = {Zhang, Kaizhong and Shasha, Dennis},
  journal      = {SIAM Journal on Computing},
  volume       = {18},
  number       = {6},
  pages        = {1245--1262},
  year         = {1989},
  doi          = {10.1137/0218082}
}

@article{levenshtein1966binary,
  title        = {Binary Codes Capable of Correcting Deletions, Insertions and Reversals},
  author       = {Levenshtein, Vladimir I.},
  journal      = {Soviet Physics Doklady},
  volume       = {10},
  number       = {8},
  pages        = {707--710},
  year         = {1966}
}

@article{robertson2009bm25,
  title        = {The Probabilistic Relevance Framework: {BM25} and Beyond},
  author       = {Robertson, Stephen and Zaragoza, Hugo},
  journal      = {Foundations and Trends in Information Retrieval},
  volume       = {3},
  number       = {4},
  pages        = {333--389},
  year         = {2009},
  doi          = {10.1561/1500000019}
}

@article{cui2026paddleocrvl15,
  title        = {{PaddleOCR-VL-1.5}: Towards a Multi-Task 0.9B {VLM} for Robust In-the-Wild Document Parsing},
  author       = {Cui, Cheng and Sun, Ting and Liang, Suyin and Gao, Tingquan and Zhang, Zelun and Liu, Jiaxuan and Wang, Xueqing and Zhou, Changda and Liu, Hongen and Lin, Manhui and Zhang, Yue and Zhang, Yubo and Liu, Yi and Yu, Dianhai and Ma, Yanjun},
  journal      = {arXiv preprint arXiv:2601.21957},
  year         = {2026}
}

@article{zhang2026paddleocrvl16,
  title        = {{PaddleOCR-VL-1.6}: Expanding the Frontier of Document Parsing with Under-Optimized Region Refinement and Progressive Post-Training},
  author       = {Zhang, Zelun and Liu, Hongen and Liang, Suyin and Zhang, Yubo and Xiang, Yiqing and Liu, Jiaxuan and Sun, Ting and Lin, Manhui and Zhang, Yue and Zhou, Changda and Gao, Tingquan and Cui, Cheng and Liu, Yi and Yu, Dianhai and Ma, Yanjun},
  journal      = {arXiv preprint arXiv:2606.03264},
  year         = {2026}
}

@article{duan2026glmocr,
  title        = {{GLM-OCR} Technical Report},
  author       = {Duan, Shuaiqi and Xue, Yadong and Wang, Weihan and Su, Zhe and Liu, Huan and Yang, Sheng and Gan, Guobing and Wang, Guo and Wang, Zihan and Yan, Shengdong and Jin, Dexin and Zhang, Yuxuan and Wen, Guohong and Wang, Yanfeng and Zhang, Yutao and Zhang, Xiaohan and Hong, Wenyi and Cen, Yukuo and Yin, Da and Chen, Bin and Yu, Wenmeng and Gu, Xiaotao and Tang, Jie},
  journal      = {arXiv preprint arXiv:2603.10910},
  year         = {2026}
}

@article{minimax2026m2,
  title        = {The {MiniMax-M2} Series: Mini Activations Unleashing Max Real-World Intelligence},
  author       = {{MiniMax Team}},
  journal      = {arXiv preprint arXiv:2605.26494},
  year         = {2026}
}

@article{glm5team2026glm5,
  title        = {{GLM-5}: From Vibe Coding to Agentic Engineering},
  author       = {{GLM-5 Team}},
  journal      = {arXiv preprint arXiv:2602.15763},
  year         = {2026}
}

@article{kimi2025k2,
  title        = {{Kimi K2}: Open Agentic Intelligence},
  author       = {{Kimi Team}},
  journal      = {arXiv preprint arXiv:2507.20534},
  year         = {2025}
}

@misc{cninfo2026,
  author       = {{Shenzhen Securities Information Co., Ltd.}},
  title        = {{CNINFO}: Listed-Company Disclosure Platform},
  year         = {2026},
  howpublished = {\url{https://www.cninfo.com.cn/}},
  note         = {Accessed July 2026}
}

@misc{sec2026edgar,
  author       = {{U.S. Securities and Exchange Commission}},
  title        = {{EDGAR}: Search Filings},
  year         = {2026},
  howpublished = {\url{https://www.sec.gov/search-filings}},
  note         = {Accessed July 2026}
}

@misc{arxiv2026,
  author       = {{arXiv}},
  title        = {About arXiv},
  year         = {2026},
  howpublished = {\url{https://info.arxiv.org/about/}},
  note         = {Accessed July 2026}
}

@misc{crossref2026,
  author       = {{Crossref}},
  title        = {Metadata Retrieval},
  year         = {2026},
  howpublished = {\url{https://www.crossref.org/documentation/retrieve-metadata/}},
  note         = {Accessed July 2026}
}

\end{document}